\documentclass[conference]{IEEEtran}
\IEEEoverridecommandlockouts
\usepackage{cite}
\usepackage{amsmath,amssymb,amsfonts}
\usepackage{algorithmic}
\usepackage{graphicx}
\usepackage{textcomp}
\usepackage[table]{xcolor}
\usepackage{booktabs}
\usepackage{multirow}
\def\BibTeX{{\rm B\kern-.05em{\sc i\kern-.025em b}\kern-.08em
    T\kern-.1667em\lower.7ex\hbox{E}\kern-.125emX}}
\begin{document}

\title{Predicting Loss Risks for B2B Tendering Processes}

\author{\IEEEauthorblockN{Eelaaf Zahid}
\IEEEauthorblockA{\textit{IBM Research - Almaden} \\
San Jose, CA, USA \\
eelaaf.zahid@ibm.com}
\and
\IEEEauthorblockN{Yuya Jeremy Ong}
\IEEEauthorblockA{\textit{IBM Research - Almaden} \\
San Jose, CA, USA \\
yuyajong@ibm.com}
\and
\IEEEauthorblockN{Aly Megahed*} \thanks{*Author performed work during employment at IBM Research.}
\IEEEauthorblockA{\textit{IBM Research - Almaden} \\
San Jose, CA, USA \\
aly300@gmail.com}
\and
\IEEEauthorblockN{Taiga Nakamura}
\IEEEauthorblockA{\textit{IBM Research - Almaden} \\
San Jose, CA, USA \\
taiga@us.ibm.com}
}

\maketitle

\thispagestyle{plain}
\pagestyle{plain}

\begin{abstract}

Sellers and executives who maintain a bidding pipeline of sales engagements with multiple clients for many opportunities significantly benefit from data-driven insight into the health of each of their bids. There are many predictive models that offer likelihood insights and win prediction modeling for these opportunities. Currently, these win prediction models are in the form of binary classification and only make a prediction for the likelihood of a win or loss. The binary formulation is unable to offer any insight as to why a particular deal might be predicted as a loss. This paper offers a multi-class classification model to predict win probability, with the three loss classes offering specific reasons as to why a loss is predicted, including no bid, customer did not pursue, and lost to competition. These classes offer an indicator of how that opportunity might be handled given the nature of the prediction. Besides offering baseline results on the multi-class classification, this paper also offers results on the model after class imbalance handling, with the results achieving a high accuracy of 85\% and an average AUC score of 0.94. 

\end{abstract}


\section{Introduction}
\label{sec:1}
For modern enterprises and organizations to be successful in driving their business objectives, many of their core information technology (IT) operations and infrastructures are often outsourced to external service providers to realize the benefits of the innovations and support systems they provide. Here, multiple service providers must compete against other providers to bid over large-scale IT service contracts through a competitive tendering process, which often comprise of complex requirements which can potentially yield high revenue if the contract has been signed \cite{greenia2014}.

\begin{figure}[ht]
    \centering
    \includegraphics[width=\columnwidth]{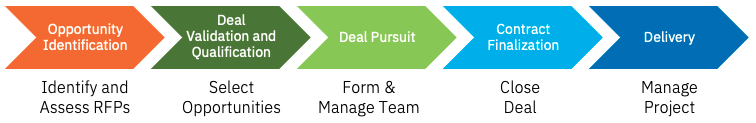}
    \caption{Contract Tendering Process Pipeline}
    \label{fig:tend_proc}
\end{figure}

Many of these service providers often bid on multiple tendering process opportunities  simultaneously, thus creating and maintaining a pipeline of sales engagements with multiple clients. This pipeline comprises of several different stages, known as \textit{sales stages}, where each bidding opportunity advances in different phases of the tendering process, as shown in Figure \ref{fig:tend_proc}. Each of these different opportunities advance through different sales stages depending on the various milestones and objectives cleared at each phase of the tendering process.

In the typical life cycle of a deal tendering process, an opportunity may fall in one of at least five different sales stages which are: 1) opportunity identification, 2) deal validation and qualification, 3) deal pursuit, 4) contract finalization, and 5) closing tendering process. 

Prior to the initialization of a tendering process, a prospective client typically distributes a Request for Proposals (RFP) to any service providers who will be taking part of the bidding process. An RFP document is a formalized documentation which is used by the client to assess the background of the service providers, their capabilities, and the potential risks involved. A service provider will then provide a response document along with other assets to present the solution they will be implementing.

In the first stage of the tendering process, the service providers seek any RFPs that are distributed by the client. In the second phase of the tendering process, the service provider analyzes the RFP to determine whether or not the corresponding deal is a suitable match against their own service offering portfolios and assess the level of trustworthiness of that client. It is also at this step that a solutioning team is assembled to piece together the preliminary solution, which is then subsequently submitted to the client. Towards the end of this phase, the client begins to filter down the number of potential service providers based on the solutions offered by each of the providers. Next, within the third phase, the provider assembles a sales team to work extensively with the client to define the scope of the solution and formulate the contract and define the contract terms of the deal. During the fourth phase of the bidding process, the client decides which service provider to work with, if any at all. If the client decides to choose a particular service provider, then the contract terms are finalized and agreed upon and the provider and client begin their work engagement to develop the solution. In the last phase, the project is handed off to the team designated to manage the project.

For sellers and executives who manage the success of the strategy and execution of these tendering processes, having a data-driven insight of the health of these individual opportunities can provide better visibility on the status of their pipeline. Furthermore, by leveraging the insight gained from knowing the likelihood that a contract will be successfully signed \cite{megahed2018}, sellers and executives can craft strategies through actionable recommendations provided by these predictive models to better navigate and optimize time, resources, and labor allocated for each of these opportunities \cite{megahed2015}.

Many predictive models that are currently in the literature mostly provide likelihood insights on the health of an opportunity \cite{megahed2015} \cite{megahed2018} \cite{greenia2014}. These models are often formulated as a binary classification model, where for each snapshot of an opportunity in the pipeline, a probability output and or a binary label (i.e., 0 or 1) is computed to determine whether a contract will be won or lost. Although these models provide an estimate for deal outcomes, these models are limited in the sense that they do not provide the scope of assessing the risk and potential root causes for why a deal may lose in particular. Losses for an opportunity may be further broken down by three common scenarios: i) the service provider not pursuing the opportunity, ii) the customer did not pursue the opportunity, and iii) the opportunity was lost to a competitor.

In the scenario where the service provider does not pursue a particular deal, possible examples of reasons why may include changes in the business objectives of the service provider or the service provider does not see an appropriate alignment between the client's need and the solutions offered. If the client decides not to pursue the opportunity, this indicates that the client did not choose any of the other service providers and decided to terminate the tendering process. This is, however, different from the other scenario where the client chooses another competitor service provider instead. From these different scenarios, having an understanding of the potential risks for pursuing such deals from an earlier time frame can help benefit providers in better assessing which opportunities in the pipeline are worth engaging in. 

In this paper, we present a novel approach to modeling the risks of an opportunity within a large IT organization's opportunity pipeline defined as a multi-class classification model which classifies the four potential scenarios for how an opportunity may win or lose as described previously. The model presented in this paper aims to help guide sellers and executives to provide a better decision support system towards making core decisions on which opportunities to keep in the pipeline and which to allocate or deallocate more resources to. 

In particular, we describe how the data was sourced, what kinds of data preprocessing methodologies were applied, and how we trained and evaluated our machine learning models. Furthermore, we also describe how we mitigated against issues of class imbalances applying a novel technique by Mahajan et. al \cite{mahajan2020}. This method utilizes a sample weight-based optimization technique to tune the relevance of each sample with respect to an objective function of interest when training the model. In our work, we evaluated our models using data from two business units and geographies and demonstrate robust average area under the curve (AUC) performance between between 0.88 and 0.94 consistently across different business unit and geography combinations.

The rest of the paper is organized as follows. We review the related work in Section \ref{sec:2}. The methodology for our modeling and evaluation process is described in Section \ref{sec:3}. We present our results, analysis and observations in Section \ref{sec:4}. Finally, we conclude our paper in Section \ref{sec:5}.

\section{Literature Review}
\label{sec:2}
In this section, we review some of the relevant prior works that have been done in the area of modeling win outcomes of tendering processes and juxtapose existing works against our proposed solution. 

Thiess et. al \cite{thiess2020} alongside other works \cite{lambert2018}, \cite{yangong2015} \cite{yanzhang2015} have proposed a machine learning model that predicts a propensity score for sales opportunities. Lambert et. al \cite{lambert2018} proposed a stacking ensemble model approach, consisting of boosting, trees, random forests and neural nets, which offers an optimal binary classification win prediction model for B2B sales. Eitle et. al \cite{eitle2019} used real data from a company's CRM system to develop a combination of three models to map the end-to-end sales pipeline process. Yin et. al \cite{yin2015} proposed a novel Bernoulli-Dirichlet predictive model that allows for monitoring the progress of current deals by predicting key milestones in an opportunity. Calixto et. al \cite{calixto2020} created a Naive Bayes model to classify sales people into specific categories. Bohanec et. al \cite{bohanec2017} attempted to explain the black box of win prediction models by offering an organizational machine learning model alongside general explanation methods to guide the decision makers in their B2B interactions. Similarly, Mahajan et. al \cite{mahajan2020} offered a framework for revenue change prediction which allows for the determination of actionable decisions in a cloud services sales application. Rezazadeh et. al \cite{rezazadeh2020} offered a workflow to improve B2B sales outcomes predictions using machine learning on a cloud-based computing platform. In a similar fashion, Megahed et. al \cite{megahed2020} documented various tools formulated to create business value at a large IT service provider, including a win prediction model that uses both structured and unstructured data. Megahed et. al \cite{megahed2015} also classified several attributes influencing the outcome of an opportunity using business insights and domain knowledge. Finally, Guo et. al \cite{guo2019} offered a machine learning model that predicts the outcome of sales opportunities using two models, logistic regression and gradient boosting. 

Although many of these models aim to provide some insight on the health and status of a particular opportunity for a pipeline through a win probability score allocated to a deal, they are all formulated on the basis of a binary classification model to determine either a win or a loss for a given opportunity. This current paper focuses on a novel machine learning model formulation that expands the binary classification win prediction model defined as a multi-class classification model. Instead of predicting a binary win or loss (or a probability value between 0 and 1) for a sales opportunity, the multi-class classification model makes a prediction from one of four classes: win, no bid, customer did not pursue, and lost to competition. The intent of this break down is to offer an explanation for why a loss might be predicted, by decomposing the loss category into three unique explanatory classes. Predictions for these four classes will allow the owner of the opportunity to better assess the potential risk of an opportunity losing. Through this insight, sellers and executives spearheading the opportunity can augment their strategy by taking actions on the troubled opportunity to potentially either increase the likelihood of the client to sign the contract or to entirely abandon the opportunity for potential time and cost savings. In the next section, we describe our methodology building a multi-class model for predicting potential outcomes of opportunities. 
\section{Methodology}
\label{sec:3}
In this section, we describe the methodology behind our proposed multi-class classification model for predicting the potential outcome of opportunities. First, we define our problem definition. Then, we elaborate on the data collection and preprocessing methodologies for our models. We describe how derived attributes are computed from the raw data which are subsequently used in the downstream transformed through additional preprocessing methods to devise our input feature for the machine learning model. We also detail the modeling process as well as some key challenges we faced due to class imbalances which emerged as a result of decomposing the loss category into smaller subsets. To mitigate this issue we utilize a sample weighted optimization-based approach to optimize sample weights against a specific metric to impose \textit{importance} for each individual sample within the training distribution of the data. Finally, we describe the experimental setup we used to evaluate our methodology and how we evaluate the models.

\subsection{Problem Definition}
In this subsection, we now concretely define some key notation, our problem definition, and objectives for the multi-class classification model. We define a set of input observation of $n$ number of snapshots of in-flight opportunities as $X \in \Bbb{R}^{n\times m}$, where $n$ is the number of samples and $m$ is the number of features pertaining to the given snapshot opportunity. Subsequently we also define our target labels of the predictive model as $Y \in \Bbb{R}^{n \times o}$, where $o$ is the number of classes in our multi-class classification model, which in this case is four, for the different win and loss scenarios described previously. The target variable $Y$ is encoded as a one-hot encoded vector which represents the presence of the particular class.

Given this, we want to train a machine learning model, $f:X_{train}\rightarrow Y_{train}$, where $f$ is a function that maps the training input features of $X_{train}$ to a specific class label output, $Y_{train}$. Hence, given a set of unseen observation, $X_{test}$, we want generate predictions using our trained machine learning model by obtaining our model as $f(X_{test}) = \hat{Y}_{test}$. Thus, our objective here is to train a machine learning model, by finding the optimal parameters such that we minimize the categorical cross-entropy loss function.

\subsection{Dataset}
In this work, we curated a dataset from one of the world's largest IT service providers in the industry which had a large centralized data-lake consisting of data from various input sources regarding the provider's sales operations. In particular, we sourced our dataset from the service provider's Customer Relationship Management (CRM) system which is used to log, monitor, and track all the historical and in-flight opportunities which are being managed by various sellers and executives across the entire organization globally. The data from this CRM system are fed into this data-lake, where various data models are used to represent different components and aspects of the opportunity pipeline, for example, the client, employees who manage the opportunity, the different services and offerings provided by the organization, the different industries which the service provider handles, and the full list of opportunities that the company has on record. Considering every opportunity has a lifetime of months to years, weekly snapshots were taken to capture the development of an opportunity at a certain time consisting of hundreds of attributes. In later sections of this paper, we provide an in-depth overview of some of the key features we used from the database to build our models.

One key data attribute of an opportunity is the \textit{sales stage}, which characterizes the different milestone phases that an in-flight opportunity goes through in the pipeline as described in Section \ref{sec:1}. In the provided CRM system, the sale stages are encoded as intermediary states between two phases (e.g., identified/verified), as opposed to discrete states (e.g., identified) as previously described. Each of these sale stage states are subsequently encoded as numerical values from 01 to 11, as shown in Table \ref{tab:sales_stage}.

\begin{table}[]
\centering
\caption{Sales Stage Code to Label Conversions}
\label{tab:sales_stage}
\begin{tabular}{@{}ccc@{}}
\toprule
\rowcolor[HTML]{FFFFFF} 
\textbf{Sales Stage Name}   & \textbf{Sales Stage Code} & \textbf{Label} \\ \midrule
Noticing                    & 01                        &                 \\ \midrule
Noticed/Identifying         & 02                        &                 \\ \midrule
Identified/Validating       & 03                        &                 \\ \midrule
Validated/Qualifying        & 04                        &                 \\ \midrule
Qualified/Gaining Agreement & 05                        &                 \\ \midrule
Cond. Agreed/Closing        & 06                        &                 \\ \midrule
Won/Implementing            & 07                        & 0               \\ \midrule
Won and Complete            & 08                        & 0               \\ \midrule
No Bid                      & 09                        & 1               \\ \midrule
Customer Did Not Pursue     & 10                        & 2               \\ \midrule
Lost to Competition         & 11                        & 3               \\ \bottomrule
\end{tabular}
\end{table}

The sales stages attribute which are coded between 01 and 06 indicate that an opportunity does not yet have an outcome. Opportunity snapshots whose states are encoded with these values are fed into the model to predict the outcome of the opportunity, such that once the model was trained, these \textit{open} opportunity snapshots are fed into the models to generate predictions on the likely outcome of the closure of the opportunity. Conversely, sales stage codes 07 to 11 indicate that an outcome was reached for the opportunity (i.e., \textit{either as won or one of the different loss states}). 

In our modeling process we utilize these codes as part of the feature of the model and also utilize these codes to encode the target label of the multi-class classification model accordingly. This means that instead of the previous win/loss prediction target discussed in a binary classification model, the multi-class target column has four unique values that are trained upon by the model. Codes 07 and 08 were labelled as 0s in the target column to indicate wins. The three reasons for loss with codes 09 through 11 were coded as 1, 2 and 3 respectively. This allows for the predictions of the model on the testing set to actually indicate the reason for a loss, as compared to just a binary win or loss representation.

\begin{table}[]
\centering
\caption{Breakdown of class counts per business unit and geographies.}
\label{tab:bu_geo_breakdown}
\begin{tabular}{cccccc}
\hline
\textbf{BU + GEO} & \textbf{Class 0} & \textbf{Class 1} & \textbf{Class 2} & \textbf{Class 3} & \textbf{Total} \\ \hline
\textbf{BU 1 GEO 1} & \begin{tabular}[c]{@{}c@{}}385058\\ (0.32)\end{tabular} & \begin{tabular}[c]{@{}c@{}}228739\\ (0.19)\end{tabular} & \begin{tabular}[c]{@{}c@{}}411654\\ (0.35)\end{tabular} & \begin{tabular}[c]{@{}c@{}}165041\\ (0.14)\end{tabular} & \textbf{1190492} \\
\textbf{BU 1 GEO 2} & \begin{tabular}[c]{@{}c@{}}492520\\ (0.40)\end{tabular} & \begin{tabular}[c]{@{}c@{}}221455\\ (0.18)\end{tabular} & \begin{tabular}[c]{@{}c@{}}378866\\ (0.31)\end{tabular} & \begin{tabular}[c]{@{}c@{}}126397\\ (0.10)\end{tabular} & \textbf{1219238} \\
\textbf{BU 1 GEO 3} & \begin{tabular}[c]{@{}c@{}}127361\\ (0.28)\end{tabular} & \begin{tabular}[c]{@{}c@{}}127258\\ (0.28)\end{tabular} & \begin{tabular}[c]{@{}c@{}}165801\\ (0.36)\end{tabular} & \begin{tabular}[c]{@{}c@{}}40434\\ (0.09)\end{tabular} & \textbf{460854} \\
\textbf{BU 1 GEO 4} & \begin{tabular}[c]{@{}c@{}}301643\\ (0.54)\end{tabular} & \begin{tabular}[c]{@{}c@{}}160635\\ (0.29)\end{tabular} & \begin{tabular}[c]{@{}c@{}}79416\\ (0.14)\end{tabular} & \begin{tabular}[c]{@{}c@{}}20041\\ (0.04)\end{tabular} & \textbf{561735} \\ \hline
\textbf{BU 2 GEO 1} & \begin{tabular}[c]{@{}c@{}}494689\\ (0.53)\end{tabular} & \begin{tabular}[c]{@{}c@{}}134388\\ (0.14)\end{tabular} & \begin{tabular}[c]{@{}c@{}}204237\\ (0.22)\end{tabular} & \begin{tabular}[c]{@{}c@{}}98321\\ (0.11)\end{tabular} & \textbf{931635} \\
\textbf{BU 2 GEO 2} & \begin{tabular}[c]{@{}c@{}}159652\\ (0.43)\end{tabular} & \begin{tabular}[c]{@{}c@{}}69059\\ (0.20)\end{tabular} & \begin{tabular}[c]{@{}c@{}}91436\\ (0.26)\end{tabular} & \begin{tabular}[c]{@{}c@{}}26903\\ (0.08)\end{tabular} & \textbf{347050} \\
\textbf{BU 2 GEO 3} & \begin{tabular}[c]{@{}c@{}}476415\\ (0.55)\end{tabular} & \begin{tabular}[c]{@{}c@{}}131897\\ (0.15)\end{tabular} & \begin{tabular}[c]{@{}c@{}}178197\\ (0.21)\end{tabular} & \begin{tabular}[c]{@{}c@{}}74489\\ (0.09)\end{tabular} & \textbf{860998} \\
\textbf{BU 2 GEO 4} & \begin{tabular}[c]{@{}c@{}}291775\\ (0.68)\end{tabular} & \begin{tabular}[c]{@{}c@{}}91126\\ (0.21)\end{tabular} & \begin{tabular}[c]{@{}c@{}}35820\\ (0.08)\end{tabular} & \begin{tabular}[c]{@{}c@{}}11291\\ (0.03)\end{tabular} & \textbf{430012} \\ \hline
\end{tabular}
\end{table}
Another key set of attributes which was integral to the dataset is the business units and the geographical location of the opportunity. As the service provider is an international organization with multiple business units and departments, the overall functions and operating characteristics of these two dimensions make a significant difference in how each business unit operates their sales strategies. To reflect this aspect of the service provider's complex organizational behavior, one key decision we applied was building many smaller individual models based on a breakdown of the different combination of business unit hierarchies and geographical regions, as opposed to having one single unified model which would be used across the entire organization.

While from a model operations and governance point of view this may pose various challenges, this discussion is beyond the scope of the paper and will not be addressed. However, there are some positive trade-offs which counters some of these main concerns. First, building smaller models allow the model to focus on a smaller subset of data distributions as opposed to a sparse distribution of attributes that spans across the organization. Further, from a scalability point of view, multiple smaller models can be trained faster and in parallel, allowing for improved run-times during deployment in an enterprise setting.

\subsection{Preprocessing}
In this subsection, we describe our preprocessing methodology for constructing final feature vectors which were utilized as part of the input for the machine learning model. In particular, we describe how to generate and encode the multi-class target labels, how we performed feature engineering through computing derived statistics from the raw data, and finally the feature selection process to identify key features to use for the model.

\subsubsection{Target Label Generation}
As previously described, the defined machine learning model generates its predictions based off of a single snapshot of an in-flight opportunity state (i.e., opportunities whose sales stage code is between 1 to 6). However, to generate the respective labels for each of these opportunity snapshots, further processing of the data is necessary to identify the end state of the opportunity's outcome in order to appropriately generate the correct ground truth labels. For this, each opportunity snapshot was grouped by its corresponding opportunity identifier and sorted by its record date. Given this, we search for the first instance of the snapshot where the sample had reached any of the closed states (i.e., sales stages 7 to 11). With this value, all opportunity snapshots where the sales stage states were considered open were allocated with this particular label value and used as the ground truth towards building the training dataset of the data. Afterwards, all subsequent opportunities where the sales stages were marked between 7 to 11 (i.e., closed states) were filtered out from the dataset entirely. The resulting preprocessed version of the label count distribution can be found in Table \ref{tab:bu_geo_breakdown}.

\subsubsection{Feature Engineering}
Besides creating the target column to hold the supervised learning label for each opportunity, a number of steps were also taken to preprocess the data. These methods include handling missing attributes for opportunities using imputation techniques, deriving new features from existing features, and ensuring the data-types of every feature correctly matched. As a result of such feature derivation, we fundamentally generated three different types of features - \textit{static features, temporal features, and derived statistics}, as summarized in Table \ref{tab:featuretypes}. Due to the sensitive nature of the data, we limit the number of features pertaining to the service provider as they are confidential.

\begin{table}[]
\centering
\caption{Types of Features}
\label{tab:featuretypes}
\begin{tabular}{@{}ccc@{}}
\toprule
\rowcolor[HTML]{FFFFFF} 
\textbf{Static Features} &
  \textbf{Temporal Features} &
  \textbf{Derived Statistics} \\ \midrule
Owning Organization &
  Sales Stage Status &
  \begin{tabular}[c]{@{}c@{}}Historical Win Rate of \\ Client (If Existing Client)\end{tabular} \\ \midrule
Geography of Client &
  \begin{tabular}[c]{@{}c@{}}Duration of Opportunity \\ Being Active (Weeks,\\  Per Sales Stages)\end{tabular} &
  \begin{tabular}[c]{@{}c@{}}Historical Win/Loss Rate \\ of Seller\end{tabular} \\ \midrule
Market of Client &
  \begin{tabular}[c]{@{}c@{}}Deal Value Fluctuation \\ Over Time\end{tabular} &
  \begin{tabular}[c]{@{}c@{}}Historical Win/Loss \\ Rate of Product\end{tabular} \\ \midrule
Deal Value &
  \begin{tabular}[c]{@{}c@{}}Update Frequency of \\ Opportunity in Records\end{tabular} &
  etc. \\ \midrule
\begin{tabular}[c]{@{}c@{}}Numer of Line \\ Items in Opportunity\end{tabular} &
  etc. &
   \\ \midrule
Comptetitors &
   &
   \\ \midrule
etc. &
   &
   \\ \bottomrule
\end{tabular}
\end{table}
Static features contain key attributes of the opportunities themselves. This includes features such as the geography of the client, the contract size of the deal, the complexity of the deal, and other features which can be directly obtained from the CRM system. Temporal features are features which compute aggregate features based on a windowed time frame, for example, computing how fast a particular opportunity moves between one sales stage to another on a weekly or quarterly basis. Derived statistics are mainly derived from core features and historical records. These are statistics which compute general percentages or rates - such as win rates for a specific department, seller or executive, or a product offering. Other features that were also used in the model include taxonomy of business units and their service offerings, client data, sponsor data and seller data. During this process, there were a total of 93 features which were generated from the raw dataset. 

After computing derived features, we further encode the raw feature vector using a categorical boosting (Catboost) encoder \cite{prokhorenkova2019} to encode the engineered features into a vector based representation before being fed into the classification model. Catboost encoders are used in the case of high cardinality features. Using a typical categorical encoding method such as one-hot encoding would lead to an extremely large number of new features. Catboost encoding groups categories first into a limited number of clusters and then applies one-hot encoding.

After feature selection, 17 features were removed due to lack of importance (as described in next subsection) and nine other features were designated as non-train features because they were either identifiers for an opportunity or correlated significantly with the target.

\subsubsection{Feature Selection}
\begin{figure}[ht]
	\centering
	\includegraphics[width=0.95\columnwidth]{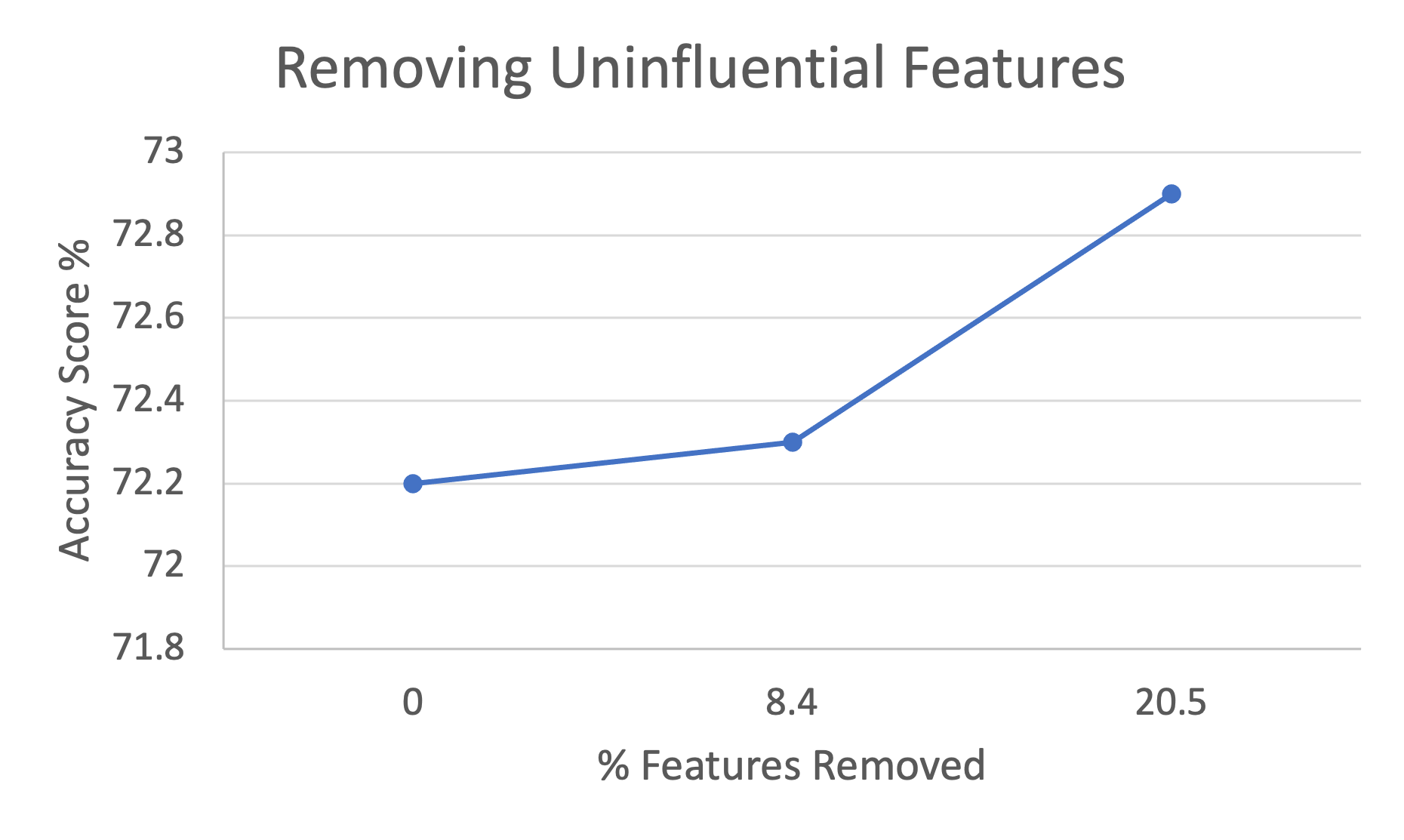}
	\caption{Feature Selection Results}
	\label{fig:featureselection}
\end{figure}

Experiments were also run performing feature selection on the set of features. The goal of feature selection was to remove the least influential attributes from being considered during training in order to not only reduce the computational cost of the model but also to improve its performance. This was done using a feature of LightGBM classifier \cite{ke2017} which outputted a ranking of the importance of each feature during the training of the model. The ranking was based on the numbers of times the feature is used in a model, with 0 indicating no impact and a higher number indicating more impact. The features with the lowest rankings were removed in increments of 10 to improve the model metrics. For example, first,  all features with a 0 importance were removed, then all features up to an importance of 10. This was done up till a ranking of 30, at which point the removal of features did not indicate an improvement in metrics. The graph in Figure \ref{fig:featureselection} below documents the feature selection experimentation which was performed on an initial BU x GEO. The graph shows the percentage of features removed during each iteration and  shows the marginal improvement in the accuracy score of the model of less than 1\% in the three iterations of the experiment.

\subsubsection{Model and Hyper-parameter Tuning}

This model used the tree-based LightGBM Classifier \cite{ke2017} to train and then make predictions. This classifier grows trees leaf wise, works well for large datasets and has many parameters that can be tuned to optimize performance.

Hyper-parameters that were tuned within the model included boosting algorithm type, class weight, number of jobs, number of leaves, minimum number of data sample in leaf, learning rate, number of iterations, and maximum bin. Originally, default values were used for all of the hyper-parameters. Attempting to optimally tune the hyper-parameters with grid search and randomized search \cite{bergstra2012} did not yield much improvement to the metrics which led us to use the same hyper-parameters for all of our experiments. 

\subsubsection{Weight Optimized Data Imbalance Handling}
One of the major challenges in building win prediction models come from the fact that historically, most opportunities in the pipeline end up being lost and only a very small handful of opportunities end up winning. This as a result creates an imbalance in the overall distribution of opportunities that are won and lost. In most prior art cases, the challenge here then becomes how to mitigate the balance between the samples when the majority of labels for the opportunities end up being biased towards losses rather than won opportunities.

However, in our work, as a result of performing the proposed multi-class label preprocessing, we now encounter the converse problem where the majority class is now the won class and the other three loss categories now become the minority. This is evident as shown in Table \ref{tab:bu_geo_breakdown}, where we demonstrate the sample count breakdown for each of the business units and geographies. In most cases, we observe that the Won classes tends to have the majority, except for BU 1 GEO 1, where Class 2 (i.e., customer did not bid) has a slightly higher ratio of samples than the won class.

To mitigate issues with class imbalances, various heuristics exist in the state of the art to address this issue. Typically these methods include under or over sampling the data set to ensure that the class distribution is equally distributed across each of the  classes, or other techniques such as Synthetic Minority-Over Technique (SMOTE) \cite{chawla2002}, which are often utilized to handle data imbalances. However, such techniques such as SMOTE are both computationally expensive and also do not generalize well for categorical attributes as well.

In our work, we apply a novel class-imbalance handling methodology based on the work of Mahajan et. al \cite{mahajan2020}, where an objective loss function is defined to utilize a meta-optimization technique to tune individual sample weight parameters against a particular user-defined metric such as F1-score, accuracy, or the area under the curve (AUC).

Given a weight value $W$, where $0 < W < 1$, we also define the set $I$ as the set of samples of all data samples in the training distribution. We also define the set $J_i$ and $J'_i$, where $i \in {0, 1, 2, 3}$, for each of the four different class indexes for our multi-class model. Thus, we can define the following:
$$I = J_0 \cup J_1 \cup J_2 \cup J_3$$

\noindent
Then the weighted loss function may be defined as follows:
$$Weighted \; Loss(W) = \sum_{i=0}^{3} \dfrac{W_{i_{k}}}{\sum_{k=0}^{3}W_{i_{k}}} Loss(X_{J_i}, Y_{J_i}, \hat{Y}_{J_i})$$

\noindent
where the $Loss(X_{J_i}, Y_{J_i}, \hat{Y}_{J_i})$ corresponds to the loss function with respect to the specific class type $J_i$. Here we also define $M(W)$, where we train a machine learning model $M$ with respect to the sample weights $W$. Given all of these definitions, our objective function can then be defined as maximizing the following function with respect to the given sample weights:

$$\max_W P_{M(W)}$$

\noindent
where P is the precision of the model metrics, or any other function, such as recall, F1-score, or AUC can be used in its place to optimize against.

\subsection{Evaluation}
In this section, we cover the methodology we used to evaluate our model performance. First describe how we backtest our models using a variant of temporal cross-fold validation. Then we discuss additional experiments we conducted on determining the best window size for the temporal cross-validation. Finally, we define the metrics for evaluating the model performance.

\subsubsection{Methodology}
The multi-class machine learning model was evaluated using a training and validation method which utilizes a temporal rolling window-based approach for splitting the train and test datasets for each fold of the data in a temporal fashion. The typical cross-validation \cite{bates2021} method is a commonly used validation method where the datasets for train and test are shuffled over k number of folds randomly, without considering the order of the data with respect to the sample dates of each of the opportunity data.

Using this approach for validating our models would not provide us with the most accurate estimate of the model performance, as this introduces a possibility of information leakage during the evaluation of the model. Hence, the temporal rolling window validation \cite{kasemsumran2006} is specifically used to ensure that data is not leaked in the train and test sets across each fold to prevent any possibility of the model gaining an unfair advantage of having access to future timestamps of the data, when it should not. Therefore it is critical that we use a rolling window-based approach to keep past and future data separate. 

\begin{figure}[ht]
    \centering
    \includegraphics[width=0.95\columnwidth]{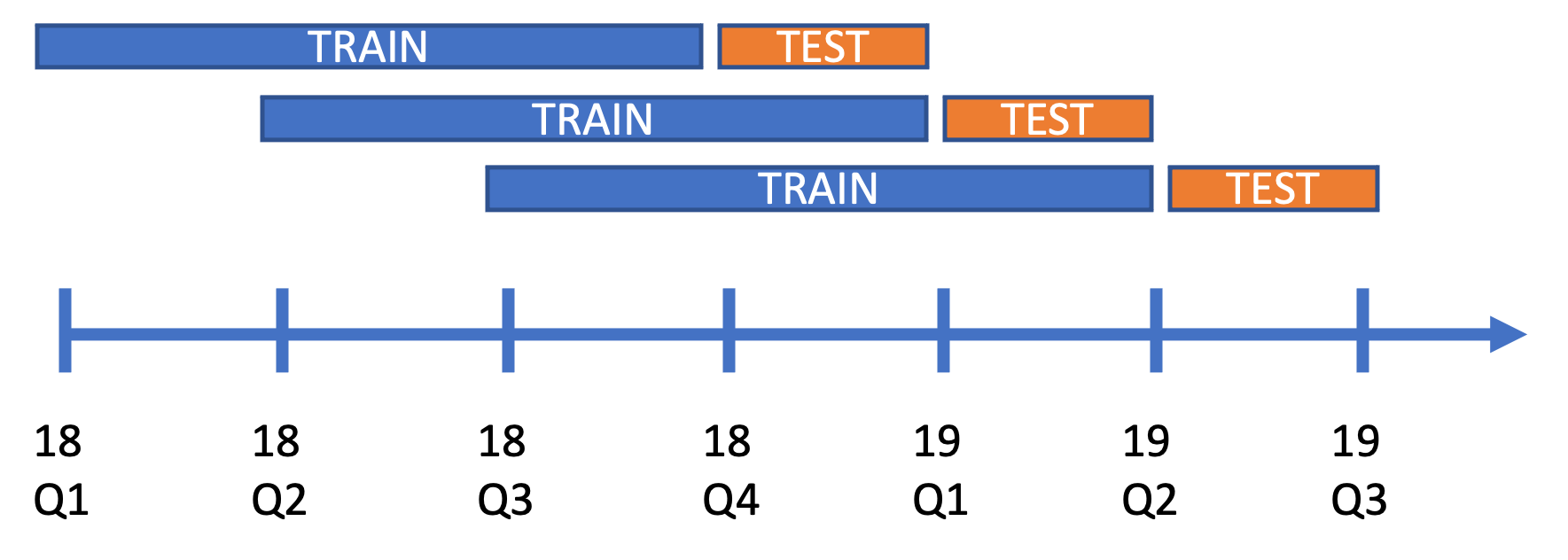}
    \caption{Rolling Window Timeline}
    \label{fig:rollingwindowtimeline}
\end{figure}

For this model evaluation methodology, rolling windows are defined as the number of quarters (in a year) used for the train and test set in one fold, with each fold removing the oldest quarter from the training set, adding the previous testing quarter as the last training quarter, and adding a new quarter for the testing set. Figure \ref{fig:rollingwindowtimeline} below offers a good visualization for this training validation method. 

As seen in Figure \ref{fig:rollingwindowtimeline}, one fold could have its training data consist of 2018Q1 - 2018Q4 and testing data from 2018Q4 - 2019Q1. Then, the next fold of validation would remove one quarter from the beginning of the training set and add one quarter at the end (which is the testing set from the last fold). This means that second fold will have a training set consisting of 2018Q2 - 2019Q1. The testing set for the second fold will shift as well so for the second fold, the testing set would be 2019Q1 - 2019Q2. This process would continue for each of the folds until there is no new quarter that can be added as the new testing set. 

The size of the total dataset and the size of the rolling window determines the number of folds training validation will have. The multi-class machine learning model had enough data that once the optimal size of the rolling window training set was set to four quarters and the testing set to one quarter, one training validation run had seven folds. In the next section, the method used to determine the optimal size of the rolling window is discussed briefly. 

\subsubsection{Rolling Window Experimentation}
\begin{figure}[ht]
    \centering
    \includegraphics[width=0.95\columnwidth]{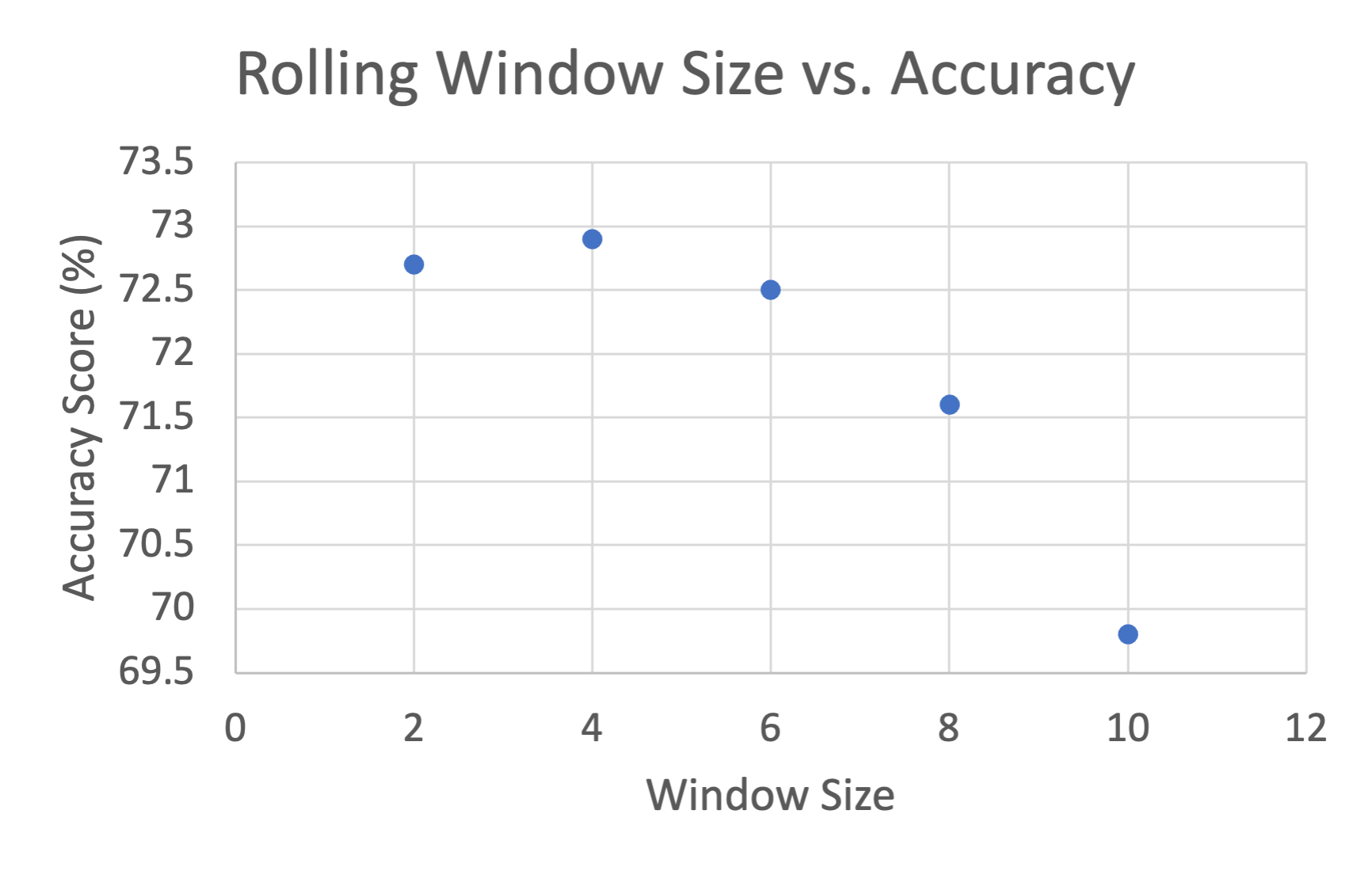}
    \caption{Rolling Window Experiment}
    \label{fig:rollingwindowgraph}
\end{figure}

During this experiment, rolling window sizes ranging from 2 through 10 were tested for the training dataset, alongside a testing window size of one quarter. For each run of the experiment, metrics were analyzed including accuracy and F1 scores for an initial BU x GEO. These metrics allowed us to determine which rolling window size was optimal for the future experiments. Based on these experiments as seen in figure \ref{fig:rollingwindowgraph}, it was determined that a rolling window size of four quarters for the training set was optimal for most experiments. 

\subsubsection{Metrics}
In our evaluation, the metrics we utilized the performance of the model included  accuracy, precision, recall, F1 score and Area Under the Curve (AUC). A typical binary classification problem only has two classes - wins and losses - and these are split into true positives (tp), true negatives (tn), false positives (fp), and false negatives (fn) when calculating metrics. Therefore, accuracy is defined as $(tp + tn)/(tp + tn + fp + fn)$. The formula for precision is $tp / tp  + fp)$. The recall for the true positive rate is $tp / (tp + fn)$ and for the false positive rate, it is $fp / (fp +tn)$. Lastly, the F1 score is calculated by $(2 * precision * recall) / precision + recall$. The results have to be in $tp, fp, tn, fn$ form in order to calculate these four metrics. 

Therefore, to provide these four metrics for the multi-class classification model, the metrics were calculated using One vs. Rest. This is done by splitting the classes into multiple binary classification problems to calculate the metrics in binary fashion. In the multi-class classification model, there are four classes - win, no bid, customer did not pursue and lost to competition - which are labelled in the target with values 0, 1, 2, and 3 respectively. To be able to calculate the metrics, one class is considered as a win and the three other class as a loss. This is then repeated with each class acting as a win and the other three as a joint loss. For example, the precision metric for class 0 would calculate correct predictions for class 0 as $tp$, correct predictions for any of the other three classes as $tn$, wrong predictions for class 0 as $fp$ and wrong predictions for the other three classes as $fn$. With this break down, a precision score for class 0 would be calculated with the formula given previously. In a similar fashion, the precision scores can be calculated for the other classes as well, along with the other metrics. 

\section{Results}
\label{sec:4}

\begin{table*}[ht]
\centering
\caption{Average metrics for each business unit, geography, and imbalance handling methods.}
\label{tab:avg_results}
\resizebox{\textwidth}{!}{%
\begin{tabular}{@{}cc|c|ccccc@{}}
\toprule
\textbf{Business Units} & \textbf{Geography} & \textbf{Imbalance Optimization} & \textbf{Avg Accuracy} & \textbf{Avg Precision} & \textbf{Avg Recall} & \textbf{Avg F1-Score} & \textbf{Avg AUC} \\ \midrule
\multirow{3}{*}{BU 1} & \multirow{3}{*}{GEO 1} & None & 0.7268 & 0.7482 & 0.7268 & 0.7316 & 0.8907 \\
 &  & Grid Search & 0.7378 & 0.7512 & 0.7378 & 0.7383 & 0.8912 \\
 &  & Bayesian Optimization & \textbf{0.7410} & \textbf{0.7554} & \textbf{0.7410} & \textbf{0.7421} & \textbf{0.8919} \\ \midrule
\multirow{3}{*}{BU 1} & \multirow{3}{*}{GEO 2} & None & 0.6475 & 0.6632 & 0.6475 & 0.6483 & 0.8493 \\
 &  & Grid Search & 0.7020 & 0.7553 & 0.7020 & 0.7103 & 0.8827 \\
 &  & Bayesian Optimization & \textbf{0.7055} & \textbf{0.7580} & \textbf{0.7055} & \textbf{0.7140} & \textbf{0.8834} \\ \midrule
\multirow{3}{*}{BU 1} & \multirow{3}{*}{GEO 3} & None & \textbf{0.6791} & \textbf{0.7572} & \textbf{0.6791} & \textbf{0.6880} & \textbf{0.8839} \\
 &  & Grid Search & 0.6513 & 0.6644 & 0.6513 & 0.6531 & 0.8513 \\
 &  & Bayesian Optimization & 0.6544 & 0.6671 & 0.6544 & 0.6557 & 0.8511 \\ \midrule
\multirow{3}{*}{BU 1} & \multirow{3}{*}{GEO 4} & None & 0.7849 & 0.7839 & 0.7849 & 0.7772 & 0.9081 \\
 &  & Grid Search & 0.7896 & 0.7864 & 0.7896 & 0.7864 & \textbf{0.9099} \\
 &  & Bayesian Optimization & \textbf{0.7918} & \textbf{0.7889} & \textbf{0.7918} & \textbf{0.7886} & 0.9093 \\ \midrule
\multirow{3}{*}{BU 2} & \multirow{3}{*}{GEO 1} & None & 0.7988 & 0.8068 & 0.7988 & 0.8000 & \textbf{0.9262} \\
 &  & Grid Search & 0.7993 & 0.8061 & 0.7993 & 0.8004 & 0.9258 \\
 &  & Bayesian Optimization & \textbf{0.8032} & \textbf{0.8078} & \textbf{0.8032} & \textbf{0.8031} & 0.9260 \\ \midrule
\multirow{3}{*}{BU 2} & \multirow{3}{*}{GEO 2} & None & 0.7296 & 0.7330 & 0.7296 & 0.7292 & 0.8900 \\
 &  & Grid Search & 0.7393 & 0.7408 & 0.7393 & 0.7366 & 0.8927 \\
 &  & Bayesian Optimization & \textbf{0.7421} & \textbf{0.7420} & \textbf{0.7421} & \textbf{0.7392} & \textbf{0.8932} \\ \midrule
\multirow{3}{*}{BU 2} & \multirow{3}{*}{GEO 3} & None & 0.7529 & 0.8136 & 0.7529 & 0.7640 & 0.9144 \\
 &  & Grid Search & 0.7653 & 0.8166 & 0.7653 & 0.7772 & 0.9116 \\
 &  & Bayesian Optimization & \textbf{0.7720} & \textbf{0.8194} & \textbf{0.7720} & \textbf{0.7827} & \textbf{0.9127} \\ \midrule
\multirow{3}{*}{BU 2} & \multirow{3}{*}{GEO 4} & None & 0.8536 & 0.8505 & 0.8536 & 0.8495 & 0.9368 \\
 &  & Grid Search & 0.8555 & 0.8529 & 0.8555 & 0.8513 & 0.9372 \\
 &  & Bayesian Optimization & \textbf{0.8581} & \textbf{0.8547} & \textbf{0.8581} & \textbf{0.8538} & \textbf{0.9377} \\ \bottomrule
\end{tabular}%
}
\end{table*}

\begin{table*}[ht!]
\centering
\caption{Class label-based break down of the AUC for each business unit, geography, and imbalance handling methods.}
\label{tab:class_results}
\resizebox{\textwidth}{!}{%
\begin{tabular}{@{}cc|c|cccc|c@{}}
\toprule
\textbf{Business Units} & \textbf{Geography} & \textbf{Imbalance Optimization} & \textbf{Class 0 AUC} & \textbf{Class 1 AUC} & \textbf{Class 2 AUC} & \textbf{Class 3 AUC} & \textbf{Avg AUC} \\ \midrule
\multirow{3}{*}{BU 1} & \multirow{3}{*}{GEO 1} & None & 0.9988 & 0.8238 & 0.8663 & 0.8738 & 0.8907 \\
 &  & Grid Search & 0.9991 & 0.8251 & 0.8646 & 0.8760 & 0.8912 \\
 &  & Bayesian Optimization & \textbf{0.9990} & \textbf{0.8265} & \textbf{0.8655} & \textbf{0.8766} & \textbf{0.8919} \\ \midrule
\multirow{3}{*}{BU 1} & \multirow{3}{*}{GEO 2} & None & \textbf{0.9987} & 0.7613 & 0.8032 & 0.8340 & 0.8493 \\
 &  & Grid Search & 0.9980 & 0.7738 & \textbf{0.8776} & 0.8815 & 0.8827 \\
 &  & Bayesian Optimization & 0.9985 & \textbf{0.7744} & \textbf{0.8776} & \textbf{0.8833} & \textbf{0.8834} \\ \midrule
\multirow{3}{*}{BU 1} & \multirow{3}{*}{GEO 3} & None & 0.9981 & \textbf{0.7778} & \textbf{0.8777} & \textbf{0.8819} & \textbf{0.8839} \\
 &  & Grid Search & \textbf{0.9987} & 0.7643 & 0.8058 & 0.8364 & 0.8513 \\
 &  & Bayesian Optimization & \textbf{0.9987} & 0.7649 & 0.8057 & 0.8351 & 0.8511 \\ \midrule
\multirow{3}{*}{BU 1} & \multirow{3}{*}{GEO 4} & None & 0.9992 & 0.8791 & 0.8779 & 0.8761 & 0.9081 \\
 &  & Grid Search & \textbf{0.9994} & \textbf{0.8807} & \textbf{0.8790} & \textbf{0.8806} & \textbf{0.9099} \\
 &  & Bayesian Optimization & 0.9993 & 0.8798 & 0.8785 & 0.8796 & 0.9093 \\ \midrule
\multirow{3}{*}{BU 2} & \multirow{3}{*}{GEO 1} & None & \textbf{0.9979} & 0.8812 & \textbf{0.9089} & 0.9168 & \textbf{0.9262} \\
 &  & Grid Search & 0.9974 & 0.8813 & 0.9073 & \textbf{0.9173} & 0.9258 \\
 &  & Bayesian Optimization & 0.9978 & \textbf{0.8815} & \textbf{0.9089} & 0.9159 & 0.9260 \\ \midrule
\multirow{3}{*}{BU 2} & \multirow{3}{*}{GEO 2} & None & \textbf{0.9985} & 0.8307 & 0.8547 & 0.8762 & 0.8900 \\
 &  & Grid Search & 0.9981 & 0.8368 & 0.8576 & 0.8784 & 0.8927 \\
 &  & Bayesian Optimization & 0.9980 & \textbf{0.8369} & \textbf{0.8593} & \textbf{0.8786} & \textbf{0.8932} \\ \midrule
\multirow{3}{*}{BU 2} & \multirow{3}{*}{GEO 3} & None & \textbf{0.9989} & \textbf{0.8539} & \textbf{0.8963} & \textbf{0.9086} & \textbf{0.9144} \\
 &  & Grid Search & \textbf{0.9989} & 0.8471 & 0.8912 & 0.9090 & 0.9116 \\
 &  & Bayesian Optimization & 0.9988 & 0.8482 & 0.8947 & 0.9091 & 0.9127 \\ \midrule
\multirow{3}{*}{BU 2} & \multirow{3}{*}{GEO 4} & None & 0.9996 & 0.9264 & 0.9037 & \textbf{0.9175} & 0.9368 \\
 &  & Grid Search & \textbf{0.9997} & 0.9276 & 0.9061 & 0.9155 & 0.9372 \\
 &  & Bayesian Optimization & 0.9996 & \textbf{0.9281} & \textbf{0.9063} & 0.9167 & \textbf{0.9377} \\ \bottomrule
\end{tabular}%
}
\end{table*}
In this section, we show our experimental results and key observations from the different experiments conducted on different business unit and geographies. The metrics from these experiments with our model and optimizations can be found in Tables \ref{tab:avg_results} and \ref{tab:class_results}. Table \ref{tab:avg_results} offers metrics for each business units and geography combination for the baseline model and imbalance optimization using Grid Search and Bayesian optimization. Table \ref{tab:class_results} showcases the AUC score breakdown for each of the four classes. 

\subsection{Experimental Analysis}
As shown in Table \ref{tab:avg_results}, the majority of results demonstrate that both of the optimization policies, the grid search and the bayesian based methods, show significant improvements over the baseline results. Furthermore, Table \ref{tab:class_results} shows that the AUC scores for Class One tended to be lower than the others. According to Table \ref{tab:bu_geo_breakdown}, class one consistently had the third most data with class three having the least. Based on these observations, this indicates that the amount of imbalance in data for each of the classes have demonstrated significantly to the results of those classes. Handling the class imbalance with bayesian optimization offered the most improvement in the average metrics to each of the business unit and geography combinations except for BU 1 GEO 3. The highest metrics appear in BU 2 GEO 4, achieving an accuracy of 86\% and an average AUC of 0.94. It can be seen in Table \ref{tab:class_results} that for BU 2 GEO 4, each of the classes scored on average around an AUC of 0.90. Further, it is surprising to note that from Table \ref{tab:bu_geo_breakdown} that BU 2 GEO 4 did not have as much data for the three loss classes as other business units and geographies and the four classes were very imbalanced as compared to other combinations. This leads us to consider that perhaps there is an optimal amount of imbalance that Bayesian Optimization and Grid Search perform best upon. There seems to be a positive correlation between the amount of imbalance and the metrics for the model, suggesting that the class imbalance optimization algorithms perform best on highly imbalanced data, and seemed to do well in balancing the classes with the least amount of data (classes one and three). 

The Bayesian Optimization results showed significant improvement from the baseline model, meaning that the weighted optimization technique to handle class imbalance was fruitful. As mentioned previously, our attempts at hyper-parameter tuning did not improve the baseline results, which told us that the main problem was with the large amounts of class imbalance. Bayesian Optimization offered a solution that resulted in significant results, especially for the business unit and geography combinations that had the highest amounts of imbalance. This suggests that in the future, when given data with high amounts of imbalance, it would be best to utilize Bayesian Optimization before attempting to tune the hyper-parameters. 

Considering that the model performed similarly for the three different experiments for each business unit and geography shows that even though our results for class zero were very high, the model was not over fitting on any data. It is also plausible that the win class had the highest results because the data across the BU x GEO combinations is more similar from experiment to experiment, as compared to the data for the loss classes. 

\subsection{Limitations and Advancements}

Given our results show consistency, this multi-classification modelling technique offers a good expansion to the binary classification win prediction models offered by each of the related works in section \ref{sec:2}. Nonetheless, this specific model still does have a number of limitations. 

First, in this paper, the generalizability of the results is limited because only two business units and geographies  combinations are observed. The process and pipeline used can and will be extended for other business units and geographies which will allow us to see how well our model performs for other datasets. Second, the feature selection and rolling window experimentation was only performed on one business units and geography combination and the results were generalized for all of the datasets. Because of this, the designated optimal rolling window size and amount of features removed might not have been optimal for all of the experiments performed, which means that there is a need to perform more comprehensive experiments in a manner that will be more generalizable for the dataset as a whole.

Further, LightGBM classifier is a tree based gradient boosting algorithm. It will be valuable to take a look at other algorithms for multi-class classification and consider stacking models to find aggregate results using more than one technique. This will allow us to compare the results of different algorithms and leverage the pros and cons of one algorithm to another. 

Lastly, the feature set that was used was based on an existing binary classification model. Therefore, it will be useful to access or derive more features from the IT providers CRM system that are more suitable for a multi-class classification model, or offer more information for each of the specific loss classes. 

\section{Conclusion}
\label{sec:5}

In this paper, we presented a novel machine learning model that performed multi-classification for win prediction. This model is an extension to existing binary classification models and allows the owners of opportunities to have a better understanding of why a loss might be predicted. This understanding comes from the loss class in a binary classification model being broken down into three unique classes: no bid, customer did not pursue, and lost to competition. 

In section \ref{sec:3}, we described the dataset our model used, the feature set and preprocessing techniques used to develop the target feature and the feature vector created for input in the model, the rolling window and feature selection experiments, metrics and methodology used for evaluation and the weighted optimization techniques used for handling class imbalance. 

In section \ref{sec:4}, we discussed the results of our experiments, indicating that Bayesian Optimization had the highest metrics in almost all of the the business unit and geography combinations. Further, we discussed the correlation between the amount of class imbalance and the resulting metrics, where the results indicated a positive correlation - more class imbalance seemed to yield better performance by Bayesian Optimization. Lastly, we discussed the limitations of our work including potential ways to improve the generalizability of the model and to take advantage of other algorithms that perform multi-class classification. 

Overall, this multi-class machine learning model was able to provide win predictions with the loss class broken down into three descriptive classes, which means that this model and technique can be used in future works to offer better descriptive explanations for predicted losses in a B2B sales pipeline and tendering processes. Although this model offers a better idea for the reason of a loss, it does not offer a technique or route to follow to change the future of an opportunity. Future work on win prediction and multi-class classification models specifically can be extended to offer specific actions to each loss prediction which would further improve the ability of the opportunity owner to fix a losing deal early on.

\end{document}